\documentclass{article}

\usepackage{arxiv}

\usepackage[utf8]{inputenc} 
\usepackage[T1]{fontenc}    
\usepackage{url}            
\usepackage{amsfonts}       
\usepackage{lipsum}
\usepackage{booktabs}

\usepackage{fancyvrb}

\usepackage{tabularx}
\usepackage{multirow}
\usepackage{amsmath,amssymb,amsfonts}
\usepackage{algorithmic}
\usepackage{graphicx}
\usepackage{textcomp}
\usepackage{xcolor}
\usepackage{lineno}

\usepackage[
bookmarks=true,
hypertexnames=true,
hyperindex=true,
hyperfootnotes=true,
linktocpage=false,
hyperfigures=true,
breaklinks=true, 
pdfborder={0 0 0},
colorlinks=true,
linkcolor=black,
unicode=true,						
citecolor=blue,
urlcolor=blue,
pdfinfo={
Title={PharmKE: Knowledge Extraction Platform for Pharmaceutical Texts using Transfer Learning},
Author={Nasi Jofche, Kostadin Mishev, Riste Stojanov, Milos Jovanovik, Dimitar Trajanov},
Keywords={Knowledge extraction, Natural language processing, Named entity recognition, Knowledge Graphs, Drugs}
}
]{hyperref}

\title{PharmKE: Knowledge Extraction Platform for Pharmaceutical Texts using Transfer Learning}

\author{
  Nasi Jofche, Kostadin Mishev, Riste Stojanov, Milos Jovanovik, Dimitar Trajanov\\
  Faculty of Computer Science and Engineering,\\
  Ss. Cyril and Methodius Univesity in Skopje, N. Macedonia\\
  \texttt{(name.surname)@finki.ukim.mk}
}

\begin{document}
\maketitle

\begin{abstract}

\textbf{Background and Objectives}: The challenge of recognizing named entities in a given text has been a very dynamic field in recent years. This is due to the advances in neural network architectures, increase of computing power and the availability of diverse labeled datasets, which deliver pre-trained, highly accurate models. These tasks are generally focused on tagging common entities, such as \textit{Person}, \textit{Organization}, \textit{Date}, \textit{Location}, etc., however, many domain-specific use-cases exist which require tagging custom entities which are not part of the pre-trained models. This can be solved by either fine-tuning the pre-trained models, or by training custom models. The main challenge lies in obtaining reliable labeled training and test datasets, and manual labeling would be a highly tedious task.

\textbf{Methods}: In this paper we present PharmKE, a text analysis platform focused on the pharmaceutical domain, which applies deep learning through several stages for thorough semantic analysis of pharmaceutical articles. It performs text classification using state-of-the-art transfer learning models, and thoroughly integrates the results obtained through a proposed methodology. The methodology is used to create accurately labeled training and test datasets, which are then used to train models for custom entity labeling tasks, centered on the pharmaceutical domain. This methodology is applied in the process of detecting \textit{Pharmaceutical Organizations} and \textit{Drugs} in texts from the pharmaceutical domain, by training models for the well-known text processing libraries: spaCy and AllenNLP.

\textbf{Results}: The obtained results are compared to the fine-tuned BERT and BioBERT models trained on the same dataset. Additionally, the PharmKE platform integrates the results obtained from named entity recognition tasks to resolve co-references of entities and analyze the semantic relations in every sentence, thus setting up a baseline for additional text analysis tasks, such as question answering and fact extraction. The recognized entities are also used to expand the knowledge graph generated by DBpedia Spotlight for a given pharmaceutical text.

\textbf{Conclusions}: PharmKE is a modular platform which incorporates state-of-the-art models for text categorization, pharmaceutical domain named entity recognition, co-reference resolution, semantic role labeling and knowledge extraction. The platform visualizes the results from the models, enabling pharmaceutical domain experts to better recognize the extracted knowledge from the input texts.

\end{abstract}



\section{Introduction}
\label{sec:introduction}

We are currently facing a situation where huge amounts of data are being generated continuously, in all aspects of our lives. The main source of this data are online social media platforms and news portals. Given their volume, it is generally hard for an individual to keep track of all the information stored within the data. Historically, whenever people do not have the capacity to finish a given task, they tend to invent tools that might help them. In this case, we want to use natural language processing (NLP) tools to perform intelligent knowledge extraction (KE) and use it to filter and receive only news which are of interest to us.

In this paper, we are particularly interested in extracting named entities from the pharmaceutical domain, namely entities which represent \textit{Pharmaceutical Organizations} and \textit{Drugs}. This NLP task is referred to as named entity recognition (NER) \cite{krishnan2005named, sang2003introduction}. It aims to detect the entities of a given type in a text corpora. NER takes a central place in many NLP systems, as a baseline task for information extraction, question answering, and more.

Our interest in this topic stems from a problem we are facing in our LinkedDrugs dataset \cite{jovanovik2017consolidating}, where the collected drug products can have active ingredients (\textit{Drug} entities) and manufacturers (\textit{Pharmaceutical Organization} entities) written in a variety of ways, depending on the data source, country of registration, language, etc. Our initial work showed promising results \cite{jofche2019neddrugs}, and we want to build on it. The ambiguity in entity naming in our drug products dataset makes the process of data analysis imprecise, thus using NER to normalize these name values for the active ingredients and manufacturers can significantly improve both the quality of the dataset, as well as the results from any analytical task on top of it.

The recent advances in neural network architectures improved NER accuracy, mainly by leveraging bidirectional long-short term memory (LSTM) networks \cite{sundermeyer2012lstm, lample2016neural}, convolutional networks \cite{chiu2016named}, and lately, transformer architectures \cite{devlin2018bert}. Many language processing libraries were made available for the public throughout the years \cite{li2018survey} from both academia and industry, equipped with highly accurate pre-trained models for extraction of common entity classes, such as \textit{Person}, \textit{Date}, \textit{Location}, \textit{Organization}, etc. However, as a given business might require detection of more specific entities in text, these models should either be fine-tuned, or trained anew with corresponding datasets for the desired entity types.

The main challenge resides in obtaining a large amount of labeled training data, which is required to train a highly accurate model. Even though multiple manually labeled, highly accurate and generic datasets exist on the Web \cite{balasuriya2009named}, their usage might not be feasible for the task at hand. Relevant data might either be unavailable on the Internet, or not feasible to be labeled manually.

As a solution to this problem, we propose a methodology that can be used to automatically create labeled datasets for custom entity types, showcased in texts from the pharmaceutical domain. In our case, this methodology is applied by tagging \textit{Pharmaceutical Organizations} in pharmacy related news. We prove that it can be extended to tagging other custom entities in different texts in the pharmaceutical domain by tagging \textit{Drug} entities as well, and assessing the obtained results. The main focus is the automatic application of common language processing tasks, such as tokenization, dealing with punctuation and stop words, lemmatization, as well as the possibility for application of custom, business case related text processing functions, like joining consecutive tokens to tag a multi-token entity, or performing text similarity computations.

The overall applicability and accuracy of this methodology is assessed by using two well-known language processing libraries, spaCy \cite{honnibal2017spacy} and AllenNLP \cite{Gardner2017AllenNLP}, which come with a pre-trained model based on convolutional layers with residual connections and a pre-trained model based on Elmo embeddings \cite{peters2018deep}, respectively. The custom trained models which are able to tag the custom entity \textit{Pharmaceutical Organization} indicate high tagging accuracy when compared to the initial pre-trained models' accuracy while tagging the more generic \textit{Organization} entity over the same testing dataset. In addition, a model trained on the same dataset by fine-tuning the state-of-the-art BERT is used for gaining a better insight over the results. Lastly, a fine-tuned BioBERT \cite{lee2019biobert}, a model based on BERT architecture and pre-trained on biomedical text corpora, is also used to better assess the results.

The thorough explanation of the methodology used to generate the labeled datasets is given in the following sections, followed by custom model training and accuracy assessment.

The extracted entities can help us filter the documents and news which mention them, but in the current era of data overflow, this is not enough. Therefore, we go one step further and integrate these results in a platform which then extracts and visualizes the knowledge related to these entities. This platform currently integrates state-of-the-art NLP models for co-reference resolution \cite{peters2018deep} and Semantic Role Labeling \cite{shi2019simple} in order to extract the context in which the entities of interest appear. This platform additionally offers convenient visualization of the obtained findings, which brings the relevant concepts closer to the people which use the platform.

This knowledge extraction process is then finalized by generating a Knowledge Graph (KG) using the Resource Description Framework (RDF) \cite{Lassila98resourcedescription} - a graph-oriented knowledge representation of the entities and their relations. This provides two main advantages: the RDF graph-data model allows seamless integration of the results from multiple knowledge extraction processes of various news sources within the platform, and at the same time links the extracted entities to their counterparts within DBpedia \cite{auer2007dbpedia} and the rest of the Linked Data on the Web \cite{bizer2008linked}. This provides the users of the platform with a uniform access over the entire knowledge extracted within the platform, and the relevant linked knowledge already present in the publicly available knowledge graphs.

\section{Related Work}
\label{sec:relatedwork}

Named entity recognition (NER), as a key component in NLP systems for annotating entities with their corresponding classes, enriches the semantic context of the words by adding hierarchical identification. Currently, there is a lot of new work being done in this field, especially in the process of neural networks optimization for label sequencing, which outperform early NER systems based on domain dictionaries, lexicons, orthographic feature extraction and semantic rules. Starting with \cite{collobert2011natural}, neural network NER systems with minimal feature engineering have become popular, due to the performances they achieve. They do so by introducing unified task-independent neural sequence labeling models, using convolutional neural networks (CNN) and n-dimensional representations of words.

Character-level models treat text as distributions over characters and they are able to generate embeddings for any string of characters within any textual context. With this, they improve the generalization of the model on both frequent and unseen words, which makes them popular in the biomedical domain. A model based on stacked bidirectional long short-term memory (LSTM) is introduced in \cite{kuru2016charner}. This model inputs characters and outputs tag probabilities for each character, achieving state-of-the-art NER performance in seven languages without using additional lexicons and hand-engineered features. In \cite{kim2016character}, authors present a language model composed of a CNN and LSTM, where they use characters as input to form a word representation for each token in the sentence, thus they outperform word/morpheme-level LSTM baselines.

In \cite{yao2015biomedical}, authors propose a Biomedical Named Entity Recognition (Bio-NER) method based on a deep neural network architecture, which leverages word representations pre-trained on unlabeled data collected from the PubMed database with a skip-gram language model. In \cite{habibi2017deep}, authors utilized word embedding techniques to capture the semantics of the words in the sentence and built a generic model based on long short-term memory network-conditional random field (LSTM-CRF), which outperforms state-of-the-art entity-specific NER tools.

Starting from 2018, Sequence-to-Sequence (Seq2Seq) architectures which work with text became a popular topic in NLP, due to their powerful ability to transform a given sequence of elements into another sequence – a concept which fits well in machine translation. Transformers are models which implement Seq2Seq architecture by using an encoder-decoder structure. 

One of the latest milestones in this development is the release of Google's BERT \cite{devlin2018bert} which is based on a transformer architecture and integrates an attention mechanism \cite{vaswani2017attention}. It produces outstanding results on many NLP tasks, including NER, due to its ability to learn contextual relations between words (or sub-words) in a text, making it applicable in the biomedical and pharmaceutical domains. Hakala and Pyysalo \cite{hakala-pyysalo-2019-biomedical} present an approach based on Conditional Random Fields (CRF) and multilingual BERT for biomedical named entity recognition on content in Spanish. In \cite{souza2019portuguese}, authors explore feature-based and fine-tuning training strategies for the BERT model for NER in Portuguese. Lamurias and Couto \cite{lamurias2019lasigebiotm} present an approach based on a transformer architecture for question answering in the biomedical domain.

BioBERT \cite{biobert2019} is a domain-specific language representation, pre-trained on large scale biomedical corpora. It is pre-trained on large general domain corpora (English Books, Wikipedia, etc.) and on biomedical domain corpora (PubMed abstracts, PMC full-text articles), using the BERT architecture. This language model provides improved results in various biomedical text mining tasks, including NER.

Transfer learning, as a machine learning method, provides the concept of re-usability in neural networks, where one model developed for a task can be reused as the starting point of the training process of another problem that has a significantly smaller training set. In recent years, transfer learning is one of the most popular approaches in computer vision and NLP tasks, since it out-performs the state-of-the-art models in many use-cases, and does so by using smaller training sets for fine-tuning and far less computational resources.

Transfer learning has enabled an increase of the F1 score for co-reference resolution tasks over the past few years, allowing it to reach a satisfying average of 73\%. This task is focused on clustering mentions within a text that refer to the same underlying real-world entities. Different approaches use biLSTM and attention mechanisms to compute span representations and then find co-reference chains through a softmax mention ranking model \cite{lee2017end}. Adding ELMO and coarse-to-fine \& second-order inference to this approach has resulted in a significant improvement over the F1 score achieving the above mentioned average of 73\%. This task is evaluated with the OntoNotes co-reference annotations from the CONLL2012 shared task \cite{pradhan2012conll}, which involved predicting co-reference in English, Chinese, and Arabic, using the final version (5.0) of the OntoNotes corpus. It provides an accurate and integrated annotation of multiple levels of the shallow semantic structure in text in multiple languages.

On the other hand, applying transfer learning to the task of semantic role labeling shows that applying a simple BERT-based model can achieve state-of-the-art performance compared to the previous state-of-the-art neural models that incorporated lexical and syntactic features, such as part-of-speech tags and dependency trees \cite{shi2019simple}. The reason lies in the fact that semantic role labeling can be decomposed into four tasks: predicate detection, predicate sense disambiguation, argument identification, and argument classification, where the predicate disambiguation task is focused on identifying the correct meaning of a predicate in a given context - allowing it to be formulated as a sequence labeling task, where BERT really shines.

There are multiple ways to construct an RDF-based Knowledge Graph (KG), which generally depend on the source data. In our case, we work with extracted and labeled data, so we can utilize existing solutions which recognize and match the entities in our data with their corresponding version in other publicly available KGs. One such tool is DBpedia Spotlight, an open source solution for automatic annotation of DBpedia entities in natural language text \cite{isem2013daiber}. It provides phrase spotting and disambiguation, i.e. entity linking, for the provided input. Its disambiguation algorithm is based upon cosine similarities and a modification of TF-IDF weights. The main phrase spotting algorithm is exact string matching, which uses LingPipe's\footnote{\url{http://alias-i.com/lingpipe}} Aho-Corasick implementation.

There are many platforms like AllenNLP \cite{Gardner2017AllenNLP} and Spacy\cite{honnibal2017spacy}, which aim to provide demo pages for NLP model testing, and code snippets for easier usage by the machine learning experts. On the other hand, projects like Hugging Face' Transformers\cite{Wolf2019HuggingFacesTS} and Deep Pavlov AI \cite{burtsev2018deeppavlov} are libraries that significantly speed up prototyping and simplify the creation of new solutions based on the existing NLP models.

However, to the best of our knowledge, there is no complete solution for knowledge extraction in the pharmaceutical domain that is human-centric and enables visualisation of the results in a human-understandable format. In this paper, we present a platform which tries to fill this gap.

\section{PharmKE Knowledge Extraction Platform}
\label{sec:application}

\begin{figure}[!ht]
    \centering
    \includegraphics[width=0.9\textwidth]{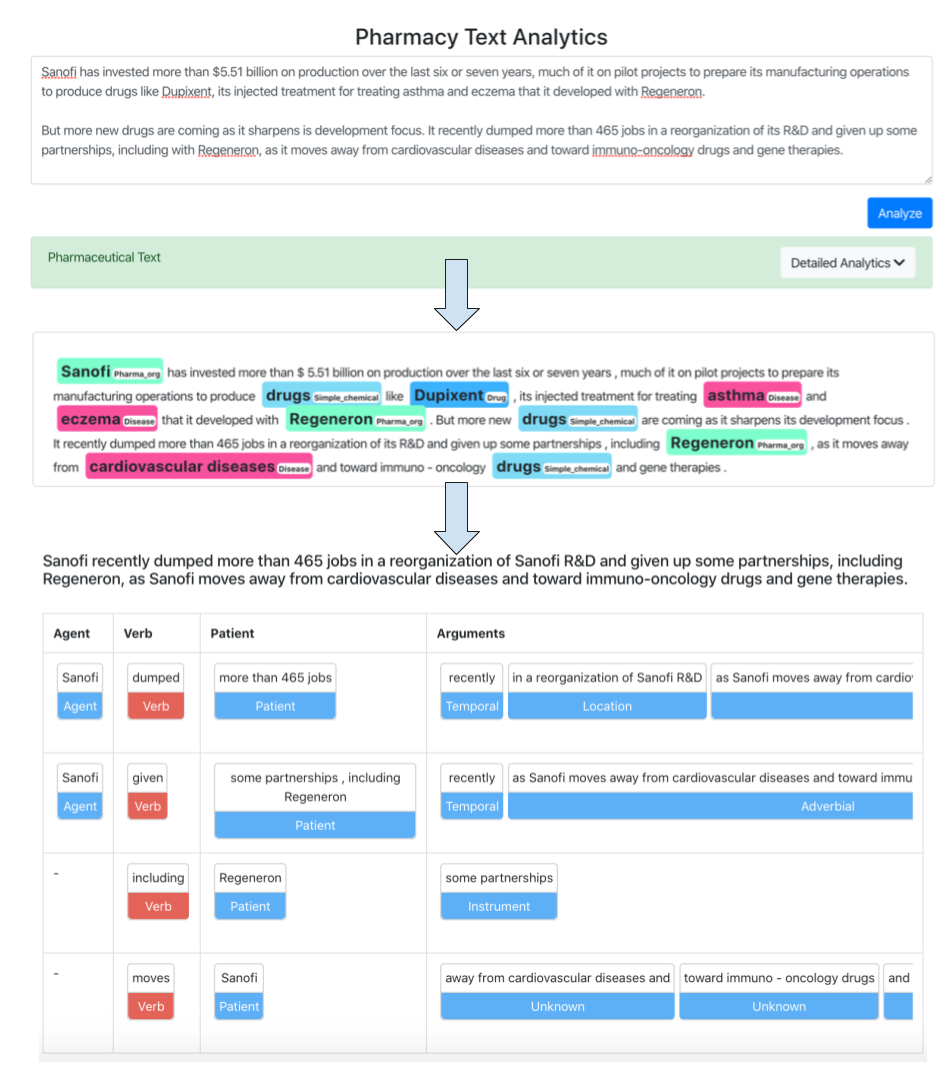}
    \caption{Platform workflow, available via the public instance of the platform \cite{pharmkeplatform2021}.}
    \label{fig:platform-workflow}
\end{figure}

This section describes our PharmKE platform \cite{pharmkeplatform2021, pharmkesourcecode2021}, which goes a step further in understanding pharmaceutical texts: on top of identifying Drugs and Pharmaceutical Organizations, it also extracts relations in the mentioned context and constructs a Knowledge Graph from them. The platform covers the entire process of understanding a document and its content - from its classification and filtering, i.e. does it belong to the pharmaceutical domain, all the way to visualization of the entities and their semantic relations, as shown in Fig. \ref{fig:platform-workflow}. Each of the steps is described in more detail within this section. 

The PharmKE platform can be formally represented with the following functional expression: 

\begin{flalign}
\label{eq:platformFn}
\includegraphics[width=0.4\linewidth]{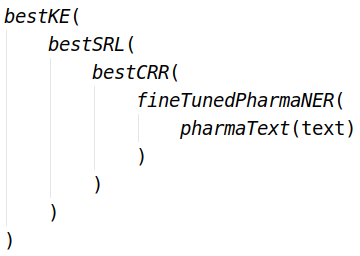}
\end{flalign}
 
The functional expression (\ref{eq:platformFn}) shows that the platform is designed to combine the best of the available models in each of the steps, while also enabling us to fine-tune some of the models, as is the case with the $fineTunedPharmaNER$ model, which is explained in more details in Section \ref{sec:methodology}. 

\subsection{Pharmaceutical Text Detection}

At the beginning, the platform classifies whether a given text is from the pharmaceutical domain, and only the positively classified texts are accepted for further analysis. The classification model used in this step is a transferred BERT model, fine-tuned with a corpus of ~5,000 documents from the pharmaceutical domain as positive samples\footnote{The documents are extracted from \url{https://www.fiercepharma.com/}, \url{https://www.pharmacist.com/} and \url{https://www.pharmaceutical-journal.com/}}, and general news documents as negative samples\footnote{\url{https://www.kaggle.com/snapcrack/all-the-news}}. 70\% of these documents are used for fine-tuning of BERT's and XLNet's models, and their precision, recall and F1 measure is evaluated with the remaining 30\% of the documents. Table \ref{tbl:pharma_classification} shows the results obtained by the fine-tuned models. 

\begin{table}[!ht]
\centering
\caption{Pharmaceutical text classification.}
\vspace{6pt}
\begin{tabular}{c|c|c|c} 
 \toprule
 \textbf{Model} & \textbf{Precision} & \textbf{Recall} & \textbf{F1} \\
 \midrule
 BERT & 0.9633 & 0.9528 & 0.9580 \\
 XLNet & 0.9983 & 0.9871 & 0.9926 \\
 \bottomrule
\end{tabular}
\label{tbl:pharma_classification}
\end{table}

\subsection{Pharmaceutical Named Entity Recognition}
\label{pharma-ner}

Each correctly classified pharmaceutical text is further analyzed by recognizing combined entities through the proposed models, as well as by using BioBERT for the detection of BC5CDR\footnote{\url{https://biocreative.bioinformatics.udel.edu/tasks/biocreative-v/track-3-cdr/}} and BioNLP13CG\footnote{\url{https://github.com/cambridgeltl/MTL-Bioinformatics-2016/tree/master/data}} tags \cite{wang2018cross}, which include Disease, Chemical, Cell, Organ, Organism, Gene, etc. Additionally, we use a fine-tuned BioBERT model in order to detect \textit{Pharmaceutical Organizations} and \textit{Drugs}, entity classes that are not covered by the standard NER tasks. We explain the fine-tuning process in more details in Section \ref{sec:methodology}. Tag collisions when combining the results from both models are avoided by applying precedence of the tags recognized by our fine-tuned model over the tags recognized by BioBERT's model (Simple Chemical). All of the recognized entities are visualized in the sentence, along with their respective tags.

\subsection{Co-reference Resolution and Semantic Role Labeling}
\label{crr-srl}

The recognized entities serve as a baseline for finding all of their mentions in the entire text, by applying co-reference resolution in the background and replacing each mention ("it", "it's", "his", etc.) with their respective entity. Libraries such as AllenNLP, StanfordNLP \cite{manning2014stanford} and NeuralCoref\footnote{\url{https://github.com/huggingface/neuralcoref}} provide implementations of the algorithms for co-reference resolution, focused on the CONLL2012 shared task \cite{pradhan2012conll}. Our platform utilizes the NeuralCoref library for co-reference resolution due to its high accuracy, ease of integration compared to StanfordNLP, and the capability to take into account user-specific information and the speakers in a conversation.

Once the mentions in the text are replaced with their respective entities, the final task includes labeling the semantic roles in each sentence. This is performed by using the BERT-based algorithm for semantic role labeling \cite{shi2019simple}. Then, the concrete arguments, like subject and object, as well as modifier arguments like temporal, location, instrument, etc. are visualized in a sequential manner for quick understanding.

The result is a modular platform for pharmaceutical text analysis, which uses existing state-of-the-art models for entity recognition, as well as fine-tuned models for recognizing custom entities like Pharmaceutical Organization and Drug. The modular design of the platform enables a combination of results from multiple models which recognize a vast range of entities. It also allows for semantic role labeling and visualization for each entity and their respective mentions in the text, by using state-of-the-art algorithms implemented by popular libraries. The entire analysis can be exported in a JSON format, allowing it to be used for additional processing such as question answering, text summarization, fact extraction, etc.

\subsection{Knowledge Graph Generation}

As a final step, we annotate the entire text using the state-of-the-art knowledge extraction system DBpedia Spotlight \cite{mendes2011dbpedia}. The obtained results are then enriched with additional RDF facts which we construct from the identified Pharmaceutical Organization and Drug entities. This enriched knowledge graph is then available for further use within or outside the platform.

\section{Entity Recognition for Pharmaceutical Organizations and Drugs}
\label{sec:methodology}

\begin{figure}[!ht]
\centering
\includegraphics[width=\textwidth]{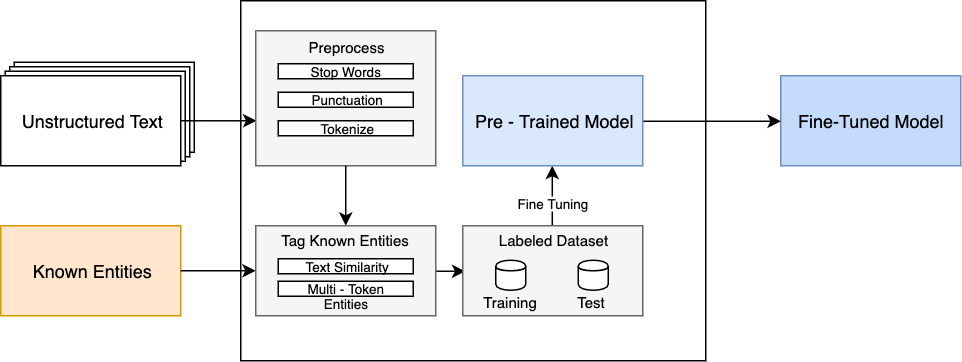}
\caption{Named entity recognition pipeline.}
\label{fig:ner-pipeline}
\end{figure}

Our methodology starts with a text corpora from the pharmaceutical domain and a closed set of entities that belong to a given class. In our case, we are using entities that denote \textit{Pharmaceutical Organizations} and \textit{Drugs}. Using only these two prerequisites, we show that we can train models that can extract even unseen entities from the class of interest. Figure \ref{fig:ner-pipeline} visualizes the whole process. 

First, we start with the text corpora from the pharmaceutical domain that potentially contains the entities from the class of interest. This text corpora consists of news collected from the following pharmacy related websites: \textit{FiercePharma}\footnote{\url{https://www.fiercepharma.com/}}, \textit{Pharmacist}\footnote{\url{https://www.pharmacist.com/}} and \textit{Pharmaceutical Journal}\footnote{\url{https://www.pharmaceutical-journal.com/}}. Next, we tokenize the text such that we extract the words, and then we try to annotate each word in respect to the set of entities from the required type. We utilize cosine similarity and levenshtein distance in particular \cite{gomaa2013survey}, where we check if the word is similar to some of the entities. The annotation process assigns start- and end-positions for each token in the text, respectively. Once we are done with this phase, we have initialized a labeled dataset, denoted as \textit{MD}. 

\subsection{Creating a Labeled Dataset}
\label{sec:creating_labeled_dataset}

One of the main challenges is that the Pharmaceutical Organization entity type can be found in a given text as multi-word phrases, such as \textbf{Sanofi Pharmaceuticals Ltd. Spain}, or as a single word: \textbf{Sanofi}. Additionally, the name of the \textit{Pharmaceutical Organization} can contain pharmacy-related keywords, such as \textbf{Pharmaceuticals}, \textbf{Pharma}, \textbf{Medical}, \textbf{Biotech}, etc., which are not part of the core name of the organization, and can either be found along with it in the sentence, or not at all. This means that we should not classify the countries, legal entities, and the pharmacy-related words as parts of the \textit{Pharmaceutical Organization} type. Therefore, the annotation process sequentially performs use-case-specific token filtering during the creation of the \textit{MD} dataset. 

This is done by using a non-entity list which contains all tokens that should be ignored. In our case, this list contains all countries in the world, together with the legal entity types for companies ("Ltd", "Inc", "GmbH", "Corp", etc.) and pharmacy-related words. After filtering out the tokens from the non-entity list, only \textbf{Sanofi} will remain in our example, and we can be certain that the core name is thoroughly extracted. After matching the core name in the text, we use the same lists to detect neighbour tokens for multi token name, if any, as part of the organization name using text similarity metrics.

After the application of the custom, use-case-related filtering, the \textit{MD} dataset consists of the core entities that have high text similarity. Only the entities which have similarity above the customized threshold are labeled as members of the target class. In our experiments, we use a similarity threshold of 0.9. Some \textit{Pharmaceutical Organization} entities consist of multiple, consecutive tokens, such as \textit{J \& J}. We solve this by token concatenation of consecutive relevant tokens, using a custom function applied on the \textit{MD}.

After applying all custom text processing functions, the state of the \textit{MD} is as shown on Table \ref{table:joined-md}.

\begin{table}[!ht]
    \caption{State of \textit{MD} after the application of the custom text processing functions.}
    \vspace{6pt}
	\centering
    \begin{tabular}{c|c|c}
     \toprule
     \textbf{Token} & \textbf{Range} & \textbf{Entity} \\
     \midrule
     Sanofi & 0:14 & PH\_ORG  \\
     GlaxoSmithKline & 258:272 & PH\_ORG  \\
     Regeneron & 3436:3440 & PH\_ORG  \\
     Regeneron & 3649:3654 & PH\_ORG  \\
     Gilead & 3660:3668 & PH\_ORG  \\
     Sanofi & 3699:3704 & PH\_ORG  \\
     J \& J & 3801:3806 & PH\_ORG  \\
     \bottomrule
    \end{tabular}
    \label{table:joined-md}
\end{table}

\subsection{Model Fine-Tuning}
\label{sec:model_fine_tuning}

The \textit{MD} dataset is then used to train a model which will be able to extract the named entities from the given class. Since NER models take into consideration the context in which the entities appear in a sentence, the training dataset is not required to contain a huge number of diverse entities. Here we improve the general knowledge language model for the more specific task, using small or moderate amounts of labeled data.

In our case, we fine-tune spaCy, AllenNLP, BERT and BioBERT models. However, each of these models requires a different data format. SpaCy requires an array of sentences with respective tagged entities for each sentence and their start- and end-positions. AllenNLP requires a dataset in BIOUL or BIO notations\footnote{\url{https://natural-language-understanding.fandom.com/wiki/Named\_entity\_recognition}}, which differentiate the following token annotations: 

\begin{itemize}
    \item multi-word entity beginning token: \textit{(B)}, 
    \item multi-word entity inside tokens: \textit{(I)} 
    \item multi-word entity ending token: \textit{(L)}, 
    \item single-token entities: \textit{(U)}, 
    \item non-entity tokens: \textit{(O)}.
\end{itemize}

The dataset adapted for BERT and BioBERT labels the entities with \textit{I - PH\_ORG}, regardless of the number of tokens, while all other tokens are marked with \textit{O}.

Therefore, we use different dataset serializers to output the training and test datasets for the fine-tuning process, in the required format.

The same methodology is used for creating labeled datasets for the \textit{Drug} entity type. In this case we use the same text corpora, but this time annotated with a fairly larger set of \textit{Drug} entities.

Once we are done with the fine-tuning process, we have named entity recognition models able to extract the entities from a given type.

\subsection{Evaluation}
\label{sec:evaluation}

The accuracy of our proposed approach is assessed by using a pharmacy-related news dataset, which consists of ~5000 news. The \textit{Pharmaceutical Organization} entities set consists of 3,633 unique values, while the \textit{Drug} entities set consists of 20,266 unique drug brand names. These sets were extracted and published as part of our previous work \cite{jovanovik2017consolidating}\cite{jofche2019neddrugs}. 

The evaluation is performed in two distinct scenarios for both entity classes. In the first evaluation, we split the news dataset into training and test portions, with sizes of 70\% and 30\% respectively, with no consideration of the distribution of the entities inside. This scenario aims to check the overall precision of the fine-tuned model. In the second evaluation scenario, we evaluate the generalization ability of our approach. Here, we split the training and test portions based on the entities they contain, such that there will not be any entity overlap between them. To do so, we extract the documents that contain 30\% of the entities as the testing portion, and the other news are used for training. However, the testing portion contained more than 30\% of the overall news. Therefore, in order to achieve a 70\% - 30\% ratio between the training and test portions, the test portion was reduced to contains exactly 30\% of the news, while in the rest of the documents, the entities were replaced with other entities which do not belong to the entity set used in the testing portion.

\subsubsection{Entity Recognition for Pharmaceutical Organizations}

\begin{table}[!ht]
    \centering
    \caption{Evaluation of models trained on a dataset that contains known entities.}
    \vspace{6pt}
    \begin{tabular}{c|c|c|c|c} 
     \toprule
     & \multicolumn{2}{|c|}{\textbf{PH\_ORG}} & \multicolumn{2}{|c}{\textbf{Organization}}  \\
     \midrule
      \textbf{Library} & \textbf{Precision} & \textbf{F1} & \textbf{Precision} & \textbf{F1} \\
     \midrule
     AllenNLP & 95.57 & 90.3 & 49.41 & 48.26 \\ 
     spaCy & 91.36 & 91.54 & 22.22 & 29.10 \\
     BERT & 97.65 & 96.66 & 51.65 & 53.18 \\
     BioBERT & 98.35* & 96.86* & 52.12* & 53.38* \\
     \bottomrule
    \end{tabular}
    \label{tbl:eval-pc}
\end{table}

\begin{table}[!ht]
    \centering
    \caption{Evaluation of the models on previously unseen entities.}
    \vspace{6pt}
    \begin{tabular}{c|c|c|c|c} 
     \toprule
     & \multicolumn{2}{|c|}{\textbf{PH\_ORG}} & \multicolumn{2}{|c}{\textbf{Organization}}  \\
     \midrule
      \textbf{Library} & \textbf{Precision} & \textbf{F1} & \textbf{Precision} & \textbf{F1} \\
     \midrule
     AllenNLP & 94.76 & 89.98 & 47.12 & 46.44 \\ 
     spaCy & 90.95 & 88.51 & 21.98 & 28.01 \\
     BERT & 97.45 & 97.68 & 51.51 & 55.68 \\
     BioBERT & 97.52* & 97.86* & 52.42* & 55.70* \\
     \bottomrule
    \end{tabular}
    \label{tbl:eval-pc-unseen}
\end{table}

The obtained fine-tuned models for detecting \textit{Pharmaceutical Organization} entities using spaCy, AllenNLP, BERT and BioBERT were tested accordingly, and the results were compared to the original models before their fine-tuning, where the task was the extraction \textit{Organization} entities. The results are given in Table \ref{tbl:eval-pc}, indicating that the fine-tuned models are able to achieve significantly higher F1 score compared to the original models. Also, we can outline that AllenNLP outperforms spaCy in this NER task, a result that can be attributed to the different neural architectures used by both libraries, while the BERT model is able to outperform both. However, the pre-trained BioBERT on biomedical text is able to slightly outperform BERT in every evaluation.

Even though the pre-trained models take into consideration the sentence context in which the entities appear, we can evaluate the fine-tuned model generalization capability by creating a test dataset that contains only entities that were not seen during the training. To achieve this, we use the joint dataset of the pharmacy-related news and generate a sample of entities in a random way to achieve a 70\% - 30\% split ratio between training and test datasets, where the test dataset contains entities not encountered in the training dataset.

\begin{figure}[!ht]
    \centering
    \includegraphics[width=0.9\textwidth]{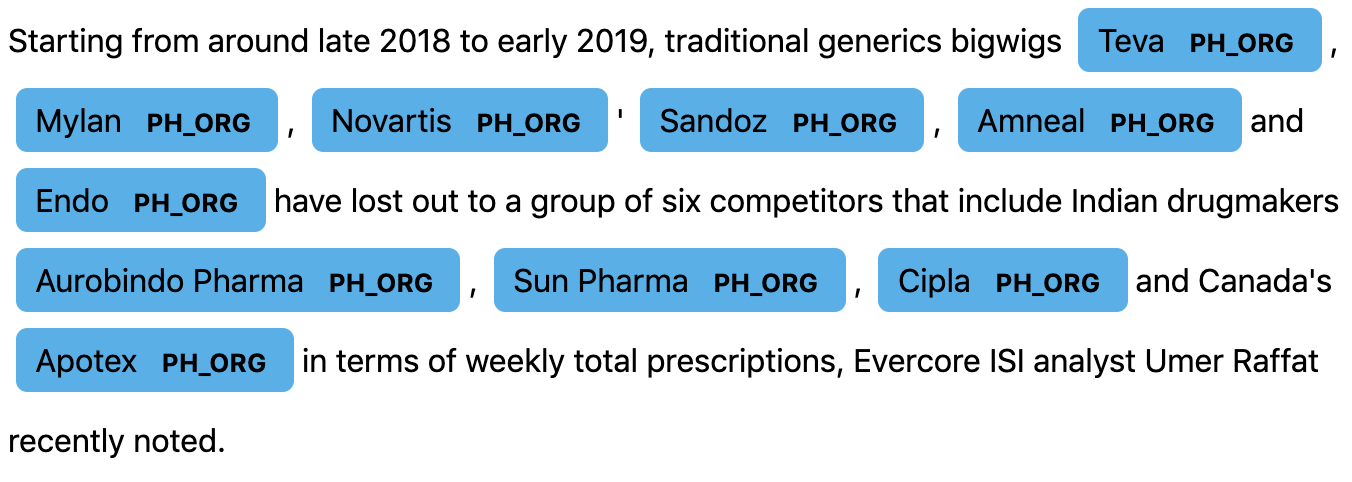}
    \caption{Detecting \textit{Pharmaceutical Organization} entities in text.}
    \label{fig:ph-org}
\end{figure}

SpaCy, AllenNLP, BERT and BioBERT models were also trained using these datasets, and the results are given in Table \ref{tbl:eval-pc-unseen}. To better visualize the accuracy, Fig. \ref{fig:ph-org} denotes a sentence extracted from pharmacy-related news where the \textit{Pharmaceutical Organization} entities are recognized as expected.

\begin{table}[!ht]
\centering
\caption{Evaluation of models trained on a dataset that contains known entities.}
\vspace{6pt}
\begin{tabular}{c|c|c} 
 \toprule
  \textbf{Library} & \textbf{Precision} & \textbf{F1} \\
 \midrule
 AllenNLP & 96.24 & 95.12 \\ 
 spaCy & 90.95 & 94.87 \\
 BERT & 98.86 & 95.98 \\
 BioBERT & 98.92* & 96.14* \\
 \bottomrule
\end{tabular}
\label{tbl:eval-drug}
\end{table}

\subsubsection{Entity Recognition for Drugs}
\label{sec:labelingdrugs}

\begin{figure}[!ht]
    \centering
    \includegraphics[width=0.8\textwidth]{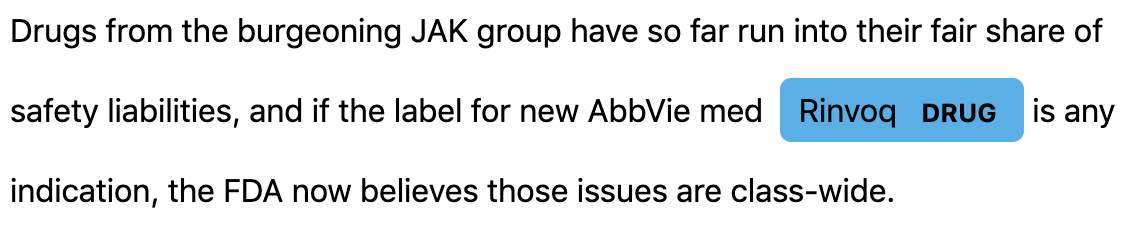}
    \caption{Detecting \textit{Drug} entities in text.}
    \label{fig:drug-example}
\end{figure}

SpaCy, AllenNLP, BERT and BioBERT models were also created for recognizing Drug entities in texts. The evaluation results are given in Table \ref{tbl:eval-drug} for the scenario where the same \textit{Drug} entity can be present in both the training and the test dataset, while Table \ref{tbl:eval-drug-unseen} shows the results when the test dataset does not contain any of the entities used in the training phase. Again, the train-test dataset ratio is 70\% - 30\%. To better visualize the accuracy, Fig. \ref{fig:drug-example} denotes a sentence extracted from pharmacy-related news, where the \textit{Drug} entity is recognized as expected.

\begin{table}[!ht]
\centering
\caption{Evaluation of the models on previously unseen entities.}
\vspace{6pt}
\begin{tabular}{c|c|c} 
 \toprule
  \textbf{Library} & \textbf{Precision} & \textbf{F1} \\
 \midrule
 AllenNLP & 92.65 & 89.85 \\ 
 spaCy & 88.16 & 89.25 \\
 BERT & 98.12 & 95.01 \\
 BioBERT & 98.65* & 95.14* \\
 \bottomrule
\end{tabular}
\label{tbl:eval-drug-unseen}
\end{table}

\section{Knowledge Graph Generation and Enrichment}
\label{sec:knowledge_expansion}

As a final step in the pipeline, we want to generate an RDF knowledge graph (KG) with the knowledge extracted from the previous steps. One way to create a general-purpose knowledge graph is to use a tool such as DBpedia Spotlight \cite{mendes2011dbpedia}, which performs recognition of interlinked entities in the DBpedia knowledge graph. So, in theory, it can be used to recognize the drugs and pharmaceutical organizations in the texts of interest, and correctly annotate them with their semantic type. However, our experiments showed that the annotated entities are of more general types, such as \verb|schema:Organization|\footnote{\url{http://schema.org/Organization}} or \verb|dbpedia:Company|\footnote{\url{http://dbpedia.org/ontology/company}}. In addition to that, most drug entities referenced by their brand names are not annotated at all. Therefore, we decided to use the results obtained so far by the pipeline described in the previous sections, to expand the knowledge graph generated by DBpedia Spotlight with specific types: \verb|schema:MedicalOrganization|\footnote{\url{http://schema.org/MedicalOrganization}} for the recognized pharmaceutical organizations, and \verb|schema:Drug|\footnote{\url{http://schema.org/Drug}}, \verb|dbpedia:Drug|\footnote{\url{http://dbpedia.org/ontology/Drug}} for the recognized drugs. 

\begin{figure}[!ht]
    \centering
    \includegraphics[width=\textwidth]{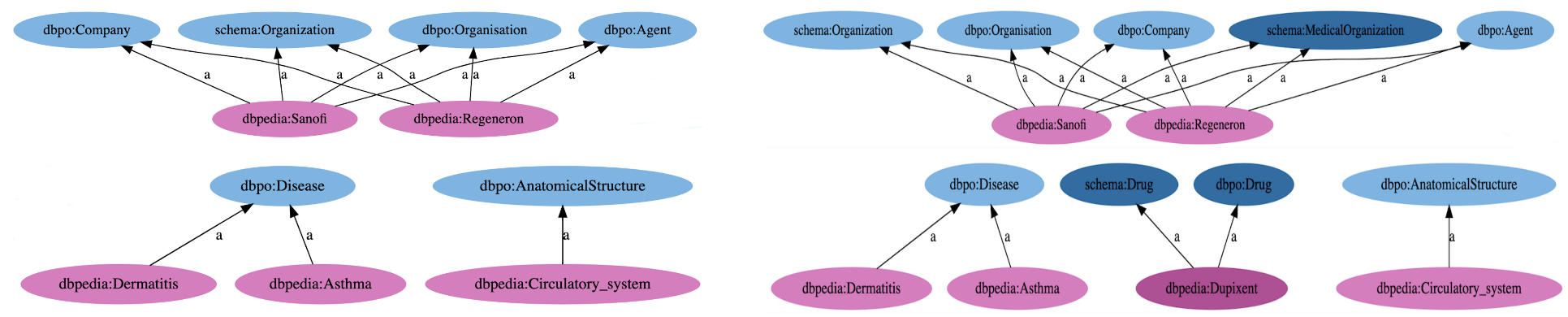}
    \caption{Original knowledge graph generated by DBpedia Spotlight (left) and the expanded knowledge graph (right). The additional RDF triples are highlighted.}
    \label{fig:knowledge-expansion-comparison}
\end{figure}

To properly test the benefits of this knowledge graph enrichment, we decided to apply the technique on the test set which contains texts with previously unknown entities while training the named entity recognition models. The results show an average expansion of 47.69\% on the originally generated knowledge graph by DBpedia Spotlight. Figure \ref{fig:knowledge-expansion-comparison} shows an example knowledge graph for a given input text, extracted using the DBpedia Spotlight annotation tool (left), and the enriched knowledge graph with additional knowledge about \textit{MedicalOrganization} and \textit{Drug} entities (right).

\begin{figure}[!ht]
    \centering
    \includegraphics[width=0.7\textwidth]{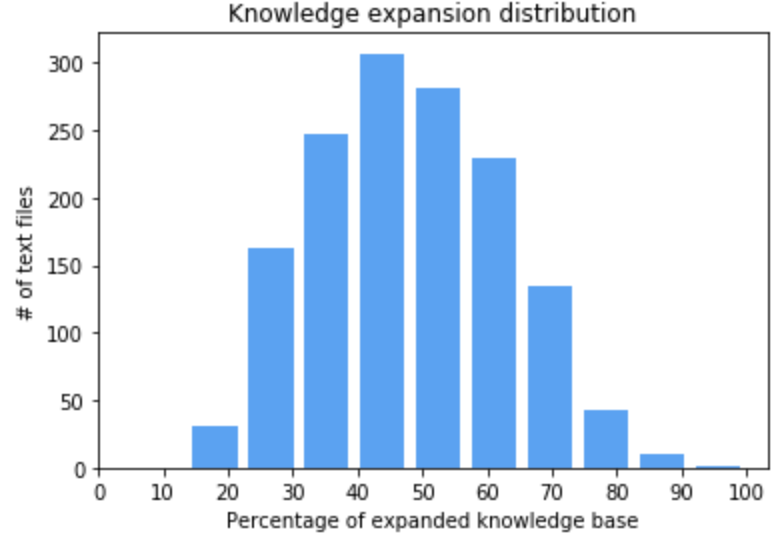}
    \caption{Distribution of knowledge graph enrichment among the texts from the test set.}
    \label{fig:files-dist}
\end{figure}

Figure \ref{fig:files-dist} shows the overall knowledge enrichment obtained by our system for the test dataset. It presents the ratio between the number of texts and the percentage of knowledge enrichment. This overview indicates a normal distribution of the enrichment over the test set.

The knowledge graph generated and enriched as part of the pipeline, can then be used for other purposes within or outside the platform. We are currently providing an RDF output in Turtle syntax\footnote{\url{https://www.w3.org/TR/turtle/}}.

\section{Discussion}
\label{sec:discussion}

The platform presented in this paper emphasizes a methodology for combining the best-performing NLP models and adopting them for use in a new domain. We use a modular approach, where each model is a separate phase in the knowledge extraction pipeline, and allows for an easy upgrade with new and potentially superior models, therefore improving the performance of the entire platform.

In contrast to \cite{Gardner2017AllenNLP}\cite{honnibal2017spacy}\cite{Wolf2019HuggingFacesTS}\cite{burtsev2018deeppavlov}, the goal of our platform is to provide a knowledge extraction solution for the pharmaceutical domain that brings the state-of-the-art NLP achievements closer to the people which analyze large amounts of texts. The PharmKE platform is human-centric, meaning that it is designed to be used primarily by people who need to extract the knowledge. The outcome from each phase is visualized, which enables the users to better understand the process of capturing and linking this knowledge. Since the web browser may not be the most convenient tool for domain experts to use in the process of knowledge extraction, especially when they analyze texts from various sources, we are also publishing an Application Programming Interface (API) that exposes the results from our platform to other applications. With this, we enable the development of editor plugins which will potentially extract and visualize the knowledge in the tools that experts already use on a daily basis.

In the current version of the PharmKE platform, we fine-tuned the Named Entity Recognition module to extract two additional entity types, namely \textit{Pharmaceutical Organization} and \textit{Drug}, on top of the entity types already recognized by the superior BioBERT model. During the fine-tuning phase, we show a method for automatically creating the training set for the recognition of \textit{Pharmaceutical Organizations} and \textit{Drugs}, by using a text corpora from the pharmaceutical domain and a closed set of entity instances from the types of interest. The evaluation of the fine-tuned model showed that this methodology enables recognition of entities that are not seen in the training set, which is a promising result.

The knowledge graph which we generate and enrich at the end of the pipeline is aimed to show the possibility of packaging and reusing the knowledge generated by the pipeline in other software solutions. Namely, even though the platform is human-centric, generating an RDF knowledge graph as the final step in the process means that the results can be stored, shared, combined with other RDF knowledge graphs and (re)used programmatically, outside of the platform. The nature of RDF and knowledge graphs allows for an almost seamless combination of the results from the platform with other RDF data which exists publicly or internally in the user environment.

The PharmKE platform is open to the continuous advancements in the NLP field. One of the crucial elements in the process of the knowledge extraction that is not solved by the current models is the linking of the relations obtained by the SRL model with the corresponding properties in the knowledge graph. This is the challenge that our team will try to address in our future research, as well as incorporating any model that will have better results in some of the current tasks. All of this is possible thanks to the modular design of the platform. Another challenge will be cleaning up the knowledge graph from erroneous conclusions made by the pipeline, which is a standard and expected problem with NLP.

\section{Conclusion}
\label{sec:conclusion}

In this paper, we present a modular platform \cite{pharmkeplatform2021, pharmkesourcecode2021} that incorporates state-of-the-art models for text categorization, pharmaceutical domain named entity recognition (NER), co-reference resolution (CRR), semantic role labeling (SRL) and knowledge extraction (KE). This platform is designed primarily for human users. PharmKE visualizes the results from each of the incorporated models, enabling pharmaceutical domain experts to better recognize the extracted knowledge from the input texts.

Our strategic goal is to keep the PharmKE platform current and up-to-date, and its modular design enables easy incorporation of new and potentially superior models. One such step in this direction was our extension of the more recent BioBERT model for NER with the \textit{Pharmaceutical Organization} and \textit{Drug} entity type recognition.

The platform is also publicly available \cite{pharmkeplatform2021} and is open-source \cite{pharmkesourcecode2021}, providing reproducibility of our results. This also means that other researchers can modify their own copy of the platform, run their own instances of it and even re-purpose it, thanks to its modular design.

A common issue while training custom models for language understanding tasks in text, is the lack of labeled datasets for testing and training. To tackle this issue, we propose a methodology that can be used to automate the labeled dataset creation process for training models for custom entity tagging. The methodology was assessed by training custom models for named entity recognition using spaCy, AllenNLP, BERT and BioBERT, and the obtained results indicate that the newly trained models outperform the pre-trained models in detecting custom entities.

\section{Future Work}
\label{sec:futurework}

Evaluating the performance of the proposed methodology on pharmaceutical texts gives satisfying results. However, a better oversight could be obtained with testing the methodology on various texts with different context, that can either include or not entities from the pharmaceutical domain. With this, we could evaluate the performance of the methodology in a generalized manner and compare the results to the current, task-specific evaluation. This would enable its usage in a variety of domains for training diverse models.

Shifting our focus towards the platform, the extracted semantic roles can be further parsed into RDF triples which comprise a knowledge graph. A platform optimization is planned as part of the future work that would enable maintenance of the knowledge graph in the background, which would be continuously enriched with every text analysis performed by the platform.

The presence of a knowledge graph in the system will enable easy access and extraction of facts by performing simple queries over the graph, and going further, it can be interconnected with other relevant knowledge graphs of the user, or public ones.

\section*{Acknowledgement}

The work in this paper was partially financed by the Faculty of Computer Science and Engineering, Ss. Cyril and Methodius University in Skopje.


\end{document}